\documentclass{article}

\PassOptionsToPackage{numbers, compress}{natbib}

\usepackage[preprint]{neur}

\usepackage[utf8]{inputenc} % allow utf-8 input
\usepackage[T1]{fontenc}    % use 8-bit T1 fonts
\usepackage{hyperref}       % hyperlinks
\usepackage{url}            % simple URL typesetting
\usepackage{booktabs}       % professional-quality tables
\usepackage{amsfonts}       % blackboard math symbols
\usepackage{nicefrac}       % compact symbols for 1/2, etc.
\usepackage{microtype}      % microtypography
\usepackage{graphicx}
\usepackage{subcaption}
\usepackage{xcolor} 
\usepackage{amsmath}
\usepackage[numbers]{natbib}  % citations

\title{KANs for Computer Vision: An Experimental Study}

\author{
  Karthik Mohan$^{1}$\thanks{This work was done during an internship at the University of Surrey.} \\
  \texttt{knm032@gmail.com} \\
  \And
  Hanxiao Wang$^{2}$ \\
  \texttt{hanxiao.wang@surrey.ac.uk} \\
  \And
  Xiatian Zhu$^{2}$ \\
  \texttt{xiatian.zhu@surrey.ac.uk} \\
  \\
  $^{1}$SASTRA University, India \\
  $^{2}$University of Surrey, United Kingdom \\
}

\begin{document}

\maketitle

\begin{abstract}
This paper presents an experimental study of Kolmogorov-Arnold Networks (KANs) applied to computer vision tasks, particularly image classification. KANs introduce learnable activation functions on edges, offering flexible non-linear transformations compared to traditional pre-fixed activation functions with specific neural work like Multi-Layer Perceptrons (MLPs) and Convolutional Neural Networks (CNNs). While KANs have shown promise mostly in simplified or small-scale datasets, their effectiveness for more complex real-world tasks such as computer vision tasks remains less explored. 
To fill this gap, this experimental study aims to provide extended observations and insights into the strengths and limitations of KANs.
We reveal that although KANs can perform well in specific vision tasks, they face significant challenges, including increased hyperparameter sensitivity and higher computational costs.  These limitations suggest that KANs require architectural adaptations, such as integration with other architectures, to be practical for large-scale vision problems. This study focuses on empirical findings rather than proposing new methods, aiming to inform future research on optimizing KANs, in particular computer vision applications or alike.

\end{abstract}

\section{Introduction}
Neural networks \cite{neural} have been pivotal in advancing computer vision, demonstrating remarkable success in processing visual data. From early applications in simple pattern recognition \cite{pattern} to sophisticated systems capable of image classification \cite{image}, object detection \cite{object}, and segmentation \cite{segment}, these architectures have become foundational in modern computer vision. Their capacity to learn complex hierarchical patterns from large data has revolutionized the landscapes of various industries \cite{industry}, driving innovations in areas such as healthcare \cite{healthcare}, security \cite{security}, and autonomous vehicles \cite{auto}.
At large, the evolution of neural network architectures in computer vision has progressed from Multi-Layer Perceptrons (MLPs)~\cite{b1,b2} to Convolutional Neural Networks (CNNs)~\cite{b3} and more recently, Vision Transformers (ViTs)~\cite{b4}. However, one of the fundamental building blocks with these deep learning networks remains unchanged -- a pre-fixed activation function (e.g. ReLU \cite{relu}, Sigmoid \cite{neural}, Swish \cite{swish}, etc) -- which brings in the critical non-linearity learning capacity as required in a wide range of target tasks.

\begin{figure}[t]
    \centering
    \begin{subfigure}[t]{0.5\textwidth}
        \centering
        \includegraphics[width=\textwidth]{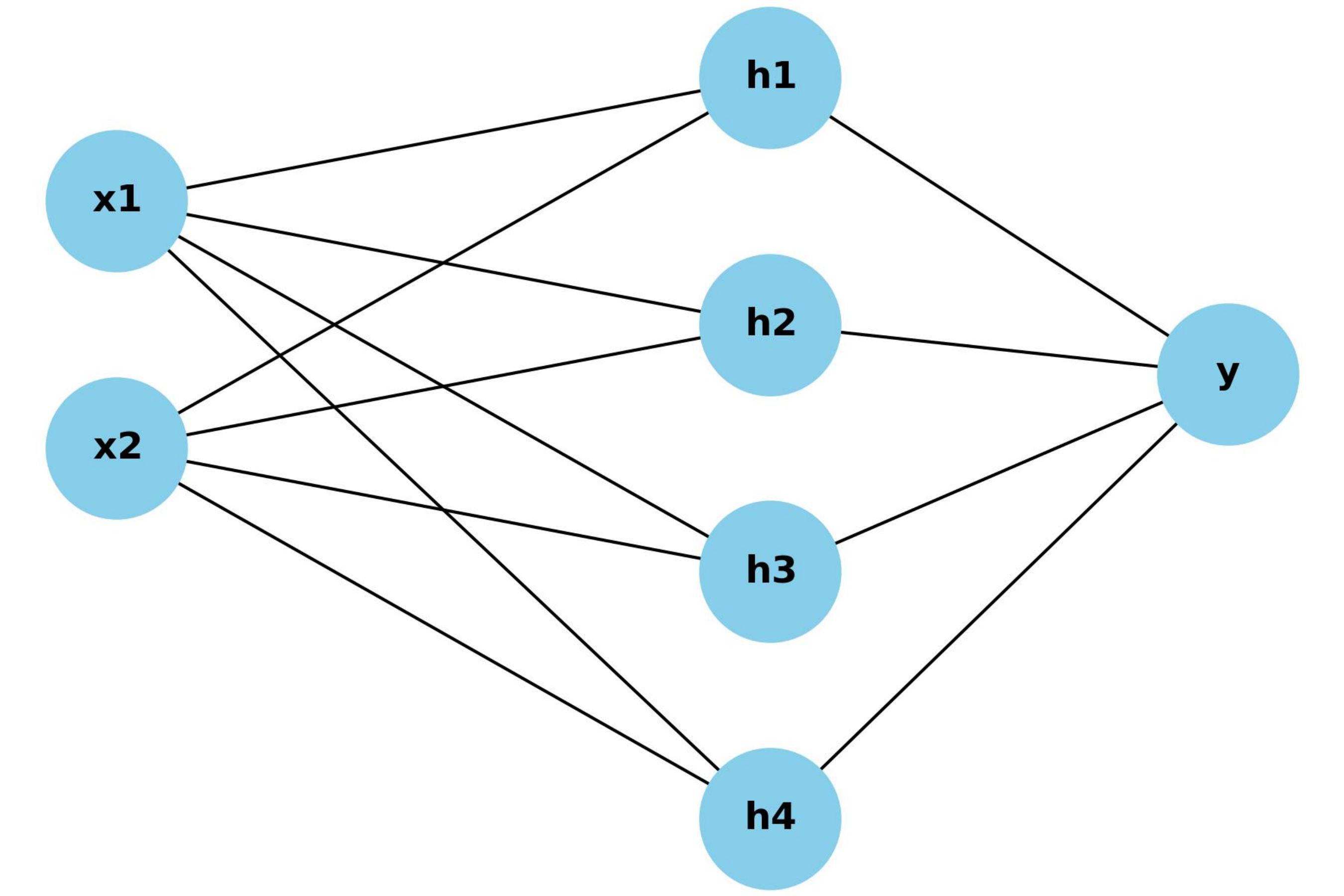}
        \caption{A standard Multi-Layer Perceptron (MLP) architecture with two input neurons, four hidden neurons, and one output neuron. Each layer applies a pre-fixed activation function, such as ReLU to the output of the neurons.}
        \label{fig:MLP}
    \end{subfigure}
    \hfill
    \begin{subfigure}[t]{0.48\textwidth}
        \centering
        \includegraphics[width=\textwidth]{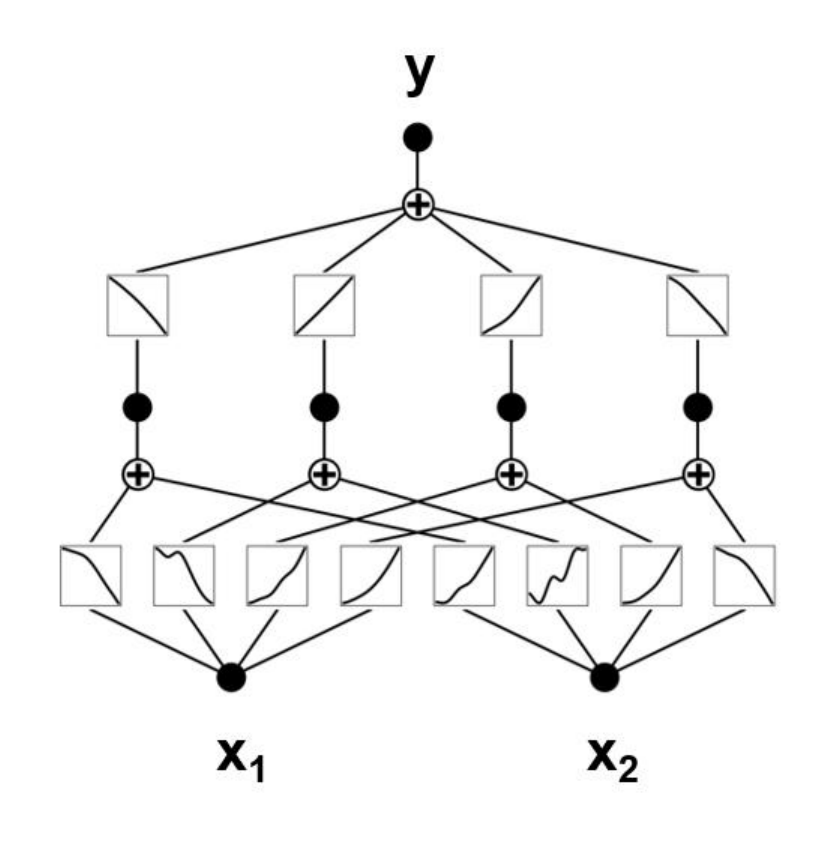}
        \caption{A KAN architecture in a $\{2,4,1\}$ structure \cite{b8}: Inputs are labeled as \( x_1 \) and \( x_2 \), and the output is \( y \). In this network, the learnable activation functions (B-splines) are placed on the edges between neurons. The edges represent the adaptable univariate functions that are learned during training.}
        \label{fig:KAN}
    \end{subfigure}
    \caption{Comparison of KAN and MLP architectures in the same model complexity.}
    \label{fig:arch_comparison}
\end{figure}

Recently a different architecture, named as Kolmogorov-Arnold Networks (KANs)~\cite{b8}, was proposed as an alternative non-linearity mechanism to the existing approaches. KANs are inspired by the Kolmogorov-Arnold Representation Theorem \cite{kar-theorem}, which states that {\em any multivariate continuous function can be decomposed into a sum of continuous univariate functions}. Unlike traditional MLPs, KANs introduce learnable activation functions on the network’s edges (rather than on nodes). These functions, typically represented by Basis splines (B-splines) \cite{bsplines}, adapt during training, offering a more flexible and potentially powerful means of modeling complex data relationships. This approach allows KANs to adapt their activation functions during training, potentially providing a more flexible and powerful modeling tool. 

However, existing research primarily focuses on symbolic regression tasks and small-scale datasets, leaving their effectiveness for large-scale tasks such as  computer vision applications largely unknown. The limited investigations into KANs have highlighted their advantages in tasks requiring precise function approximations but raised questions about their scalability and performance on larger datasets typical in computer vision.
Concurrent research has explored hybrid models that combine KANs with architectures like CNNs, aiming to integrate adaptive function representation with established convolutional feature extraction methods \cite{cnnfeature}. However, the effectiveness of such integrations in practical scenarios requires further investigation.

% Kolmogorov-Arnold Networks (KANs) \cite{b8} represents a recent and novel development in neural network architecture, inspired by the Kolmogorov-Arnold Representation Theorem, which states that any multivariate continuous function can be decomposed into a sum of continuous univariate functions. Unlike traditional MLPs, which use fixed activation functions like ReLU or Sigmoid, KANs introduce learnable activation functions on the network’s edges (rather than on nodes). These functions, typically represented by B-splines, adapt during training, offering a more flexible and potentially powerful means of modeling complex data relationships. This approach allows KANs to adapt their activation functions during training, potentially providing a more flexible and powerful modeling tool. 
% However, existing research primarily focuses on symbolic regression tasks and small-scale datasets, leaving their effectiveness in computer vision applications largely unexplored. The limited investigations into KANs have highlighted their advantages in tasks requiring precise function approximations but raised questions about their scalability and performance on larger datasets typical in computer vision.
% Recent research has explored hybrid models that combine KANs with architectures like CNNs, aiming to integrate adaptive function representation with established convolutional feature extraction methods. However, the effectiveness of such integrations in practical scenarios requires further investigation. 

This paper will address the gap by systematically evaluating the performance of KANs on classical computer vision datasets, such as MNIST \cite{mnist} and CIFAR-10 \cite{cifar}. Additionally, we explore the integration of KANs with CNNs to assess practical improvements and challenges in vision applications. We conducted extensive experiments focusing on the performance-to-parameter trade-off for the comparison between KANs and MLPs. Our experiments revealed that in our test setting KANs achieve only slightly better or equivalent performance compared to MLPs while suffering from a significantly higher number of parameters, both when applied alone or combined with other architectures like CNNs. Moreover, we found that KANs are more sensitive to additional hyperparameters (e.g., order and grid) compared to MLPs. These findings suggest that in their current form, KANs may not be a good alternative to MLPs
as part of computer vision architectures.

\section{Kolmogorov-Arnold Networks (KANs)}

In this section, we review the architectural and mathematical distinctions between Multi-Layer Perceptrons (MLPs) and Kolmogorov-Arnold Networks (KANs) \cite{b8}. We explore how these differences impact each model's expressiveness, flexibility, and computational efficiency. Additionally, we consider the potential advantages KANs might offer in specific scenarios, along with their associated challenges.

\subsection{Architecture Differences: MLP vs KAN}

Multi-layer perceptrons (MLPs) are a foundational form of feedforward neural network, structured with multiple layers of nodes, or neurons (Figure \ref{fig:MLP}). Each neuron in an MLP performs a two-step operation: a linear transformation followed by a non-linear activation. The overall transformation applied by a layer \( l \) of an MLP can be mathematically expressed as:

\begin{equation}
\mathbf{h}^{(l)} = \sigma\left(\mathbf{W}^{(l)} \mathbf{h}^{(l-1)} + \mathbf{b}^{(l)}\right)
\end{equation}

where:
\begin{itemize}
    \item \( \mathbf{h}^{(l)} \) is the output of the layer \( l \),
    \item \( \mathbf{W}^{(l)} \) is the weight matrix for the layer \( l \), representing the linear transformation,
    \item \( \mathbf{b}^{(l)} \) is the bias vector for layer \( l \),
    \item \( \sigma \) is a pre-fixed non-linear activation function (e.g., ReLU, Sigmoid, Tanh),
    \item \( \mathbf{h}^{(l-1)} \) is the input to the layer \( l \), which is the output of the previous layer \( l-1 \).
\end{itemize}

The fixed activation function \( \sigma \) introduces non-linearity into the model, enabling MLPs to approximate complex, non-linear functions. However, the expressiveness of an MLP is inherently limited by the choice of \( \sigma \), which remains constant throughout the training process. This rigidity can restrict the model’s ability to adapt to diverse data distributions or capture intricate relationships that might require more flexible transformations.

MLPs are fully connected, meaning every neuron in one layer is connected to every neuron in the subsequent layer. While this architecture allows MLPs to learn complex patterns, it also leads to a rapid increase in the number of parameters, particularly when dealing with high-dimensional input data, such as images. This growth in parameters can result in significant computational and memory demands, making MLPs less efficient for tasks involving large-scale visual data.

In contrast to MLPs, KANs feature learnable activation functions that are associated with the connections (edges) between neurons, rather than being fixed at the nodes, representing a fundamental shift architecturally (Figure \ref{fig:KAN}).

The transformation applied by layer \( l \) in a KAN is mathematically described as:

\begin{equation}
\mathbf{h}^{(l)}_i = \sum_{j=1}^{n} f_{ij}^{(l)}\left(h_j^{(l-1)}\right)
\end{equation}

where:
\begin{itemize}
    \item \( h_j^{(l-1)} \) is the \( j \)-th input to layer \( l \), inherited from the output of the previous layer,
    \item \( f_{ij}^{(l)} \) is a learnable univariate function applied to \( h_j^{(l-1)} \), typically represented by a B-spline,
    \item The sum aggregates the contributions from all inputs \( h_j^{(l-1)} \) to produce the output \( h_i^{(l)} \) for the \( i \)-th node in layer \( l \).
\end{itemize}

Unlike MLPs, where the activation function \( \sigma \) is predefined and fixed, KANs use learnable univariate functions \( f_{ij}^{(l)} \) that can be tailored to the specific data during training. These functions are often parameterized as B-splines, which are piecewise polynomial functions defined by a set of control points.

\subsection{Additional Hyperparameters in KANs}

KANs introduce two additional hyperparameters that govern the complexity and flexibility of the univariate functions: grid and order.

\begin{itemize}
    \item \textbf{Grid}: The grid refers to the number of intervals into which the input space of the univariate function is divided. For a given grid \( g \), the function is defined over \( g+1 \) control points, which determine the piecewise segments of the B-spline. A finer grid allows the function to capture more detailed variations in the data, enhancing the model's ability to fit complex patterns.

    \item \textbf{Order}: The order of the spline refers to the degree of the polynomial used in each piecewise segment. For example, \( k=2 \) corresponds to a quadratic spline, and \( k=3 \) corresponds to a cubic spline. Higher-order splines provide smoother and more flexible functions, allowing KANs to model intricate relationships with greater precision.
\end{itemize}

These hyperparameters provide KANs with a high degree of adaptability, enabling the network to adjust not only the weights but also the form of the non-linearities as required by the data. This adaptability is particularly advantageous for tasks that involve complex, non-standard data distributions, where fixed activation functions might fall short.

\subsection{Advantages and Challenges}

KANs have demonstrated superior performance in specific tasks, such as symbolic regression, where the goal is to discover underlying mathematical expressions from data \cite{superior}. This advantage can be attributed to several key factors. Firstly, the learnable activation functions in KANs provide adaptive non-linearity, enabling the model to dynamically adjust its non-linear transformations to better capture the underlying patterns in the data. This adaptability is particularly beneficial for tasks requiring precise and complex function approximations. \cite{adv} Additionally, KANs are grounded in strong mathematical principles, specifically the Kolmogorov-Arnold theorem \cite{kar-theorem}, which supports their ability to represent any continuous multivariate function as a combination of univariate functions. This makes KANs particularly well-suited to tasks that involve complex compositional structures.

Moreover, KANs offer parameter efficiency by replacing large weight matrices with smaller, learnable univariate functions. This efficiency can result in models that generalize better, especially on smaller datasets where overfitting is a concern \cite{b8}. Finally, the use of smooth splines in KANs helps to enforce smoothness in the learned functions, which is advantageous in scenarios where overfitting to noisy data could obscure the true underlying relationships. These characteristics collectively contribute to the effectiveness of KANs in tasks that demand high precision such as symbolic regression.

However, KANs also present challenges, such as their high sensitivity to hyperparameters such as grid and order \cite{hyp}.  The grid determines the number of control points for the spline functions, while the order specifies the polynomial degree of each segment. Larger values for these hyperparameters significantly amplify the parameter count, increasing the model's complexity and resource requirements. Although lower values can mitigate these demands, achieving better performance often necessitates higher values, leading to greater computational overhead. This added complexity, coupled with the requirement to tune the grid and order effectively, makes the training process more challenging compared to traditional MLPs. Improper hyperparameter choices not only hinder performance but can also result in unstable training or convergence issues, further complicating the optimization of KANs. Moreover, the added complexity of learning the activation functions, along with the weights, can make the training process more challenging and less stable compared to traditional MLPs \cite{survey}. In addition to these difficulties, another drawback of KANs is their performance in continual learning tasks. Despite claims in the original paper that KANs outperform MLPs in this setting, the work done in \cite{superior} shows that, in a standard class-incremental continual learning scenario, KANs suffer from greater catastrophic forgetting compared to MLPs

These factors must be carefully considered when choosing between KANs and MLPs for specific tasks, especially in domains like computer vision, where both performance and computational efficiency are critical.

\section{Experiments}

\subsection{Datasets}
In this study, we use three benchmark datasets commonly used in computer vision tasks: MNIST \cite{mnist}, CIFAR-10 \cite{cifar}, and Fashion-MNIST \cite{fmnist}.

\begin{itemize}
    \item \textbf{MNIST:} The MNIST dataset consists of 70,000 grayscale images of handwritten digits, each of which is 28x28 pixels in size. The dataset is divided into 60,000 training images and 10,000 test images, covering digits from 0 to 9.
    
    \item \textbf{CIFAR-10:} The CIFAR-10 dataset contains 60,000 32x32 color images across 10 different classes, such as airplanes, cars, and birds. There are 50,000 training images and 10,000 test images, with each class having 6,000 images.
    
    \item \textbf{Fashion-MNIST:} Fashion-MNIST is a dataset of 70,000 grayscale images of fashion items such as shirts, shoes, and bags. Each image is 28x28 pixels, similar to MNIST, with 60,000 images for training and 10,000 for testing.
\end{itemize}

For all datasets, the input images are provided as flattened 1D tensors, which are necessary for feeding into the Kolmogorov-Arnold Networks (KANs) and Multi-layer Perceptron (MLP) models. 
\begin{figure}[htbp]
    \centering
    \begin{subfigure}[b]{0.3\textwidth}
        \centering
        \includegraphics[width=\textwidth]{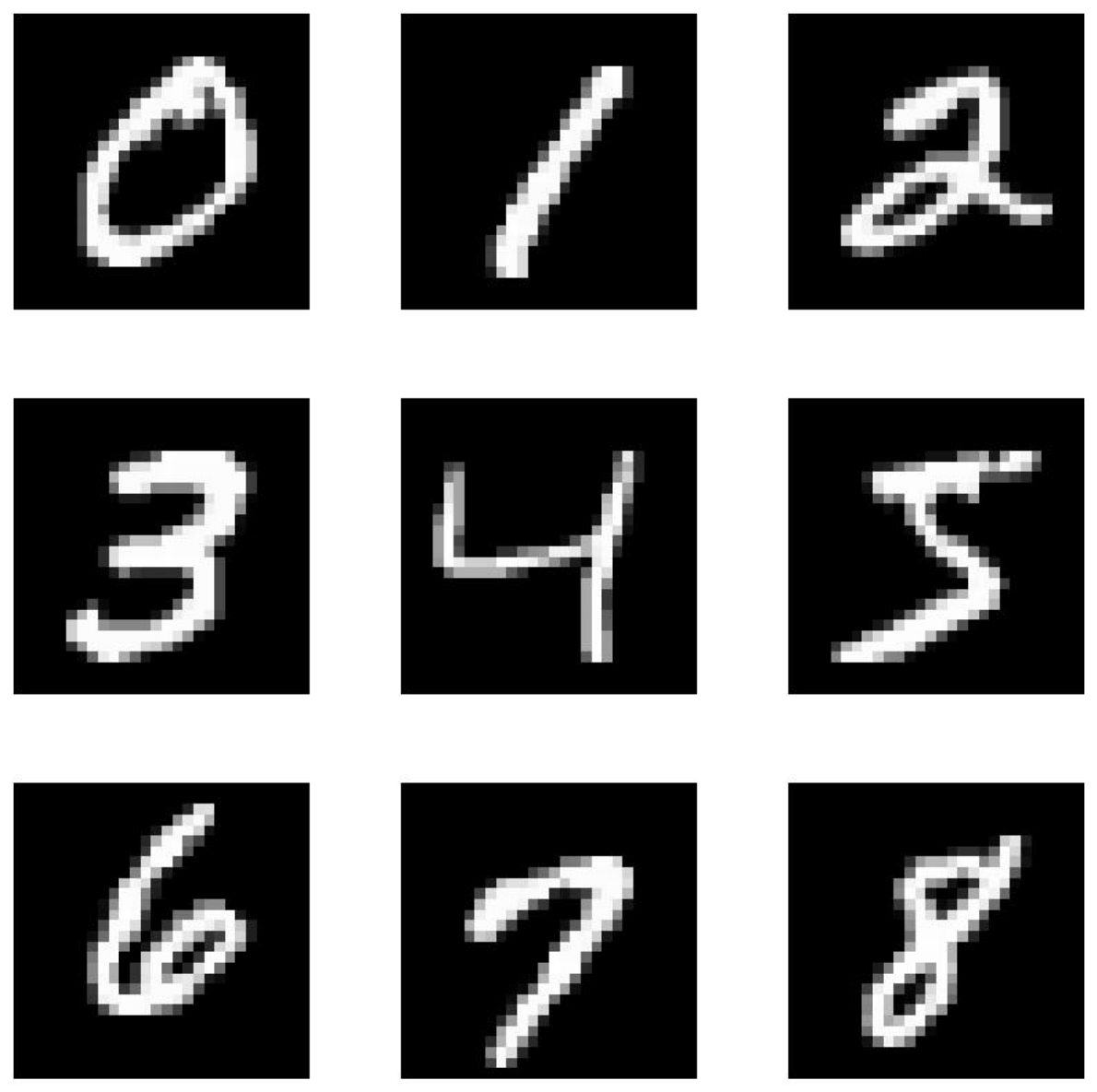}
        \caption{MNIST}
        \label{fig:mnist_sample}
    \end{subfigure}
    \hfill
    \begin{subfigure}[b]{0.3\textwidth}
        \centering
        \includegraphics[width=\textwidth]{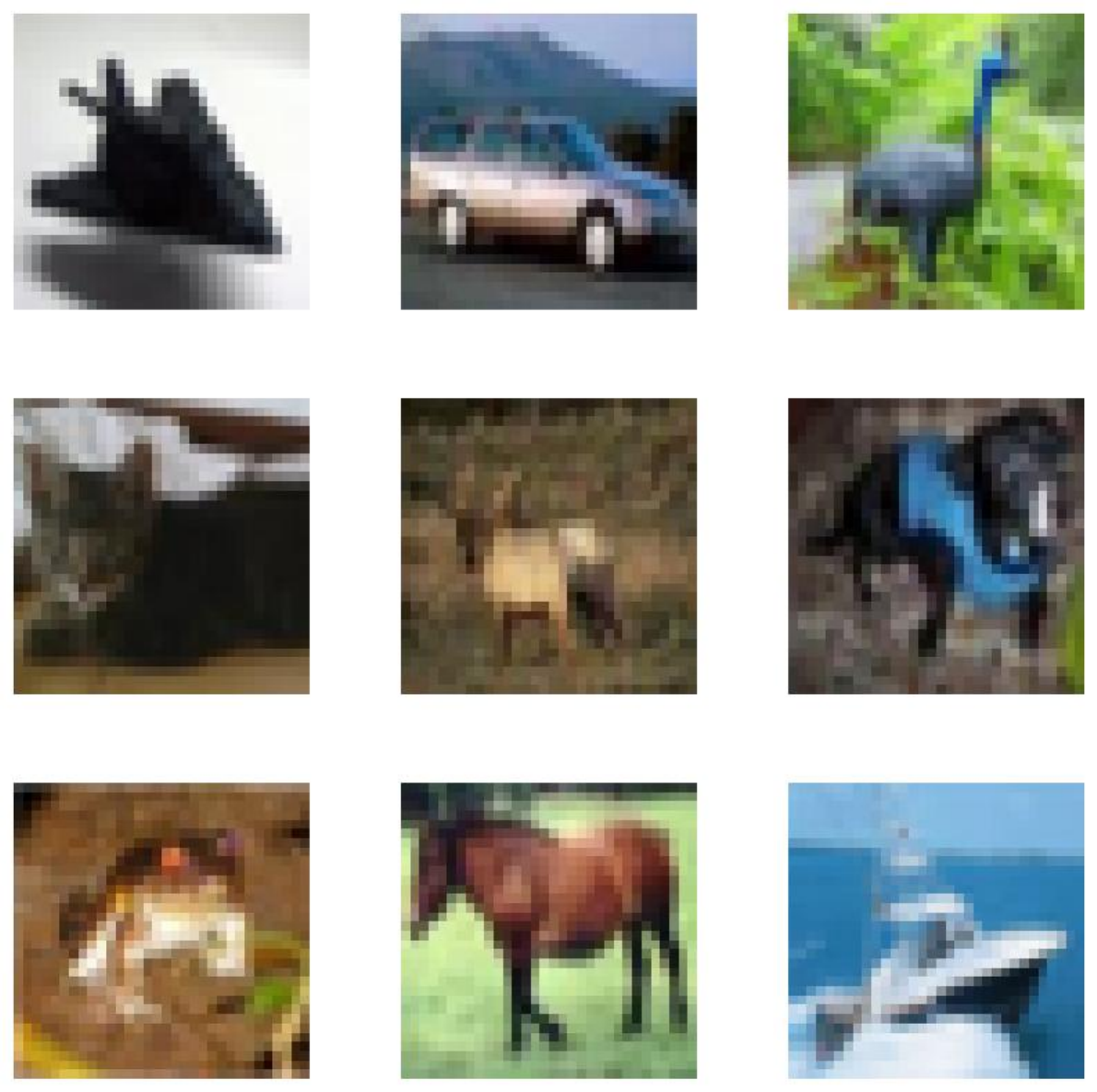}
        \caption{CIFAR-10}
        \label{fig:cifar_sample}
    \end{subfigure}
    \hfill
    \begin{subfigure}[b]{0.3\textwidth}
        \centering
        
        \includegraphics[width=\textwidth]{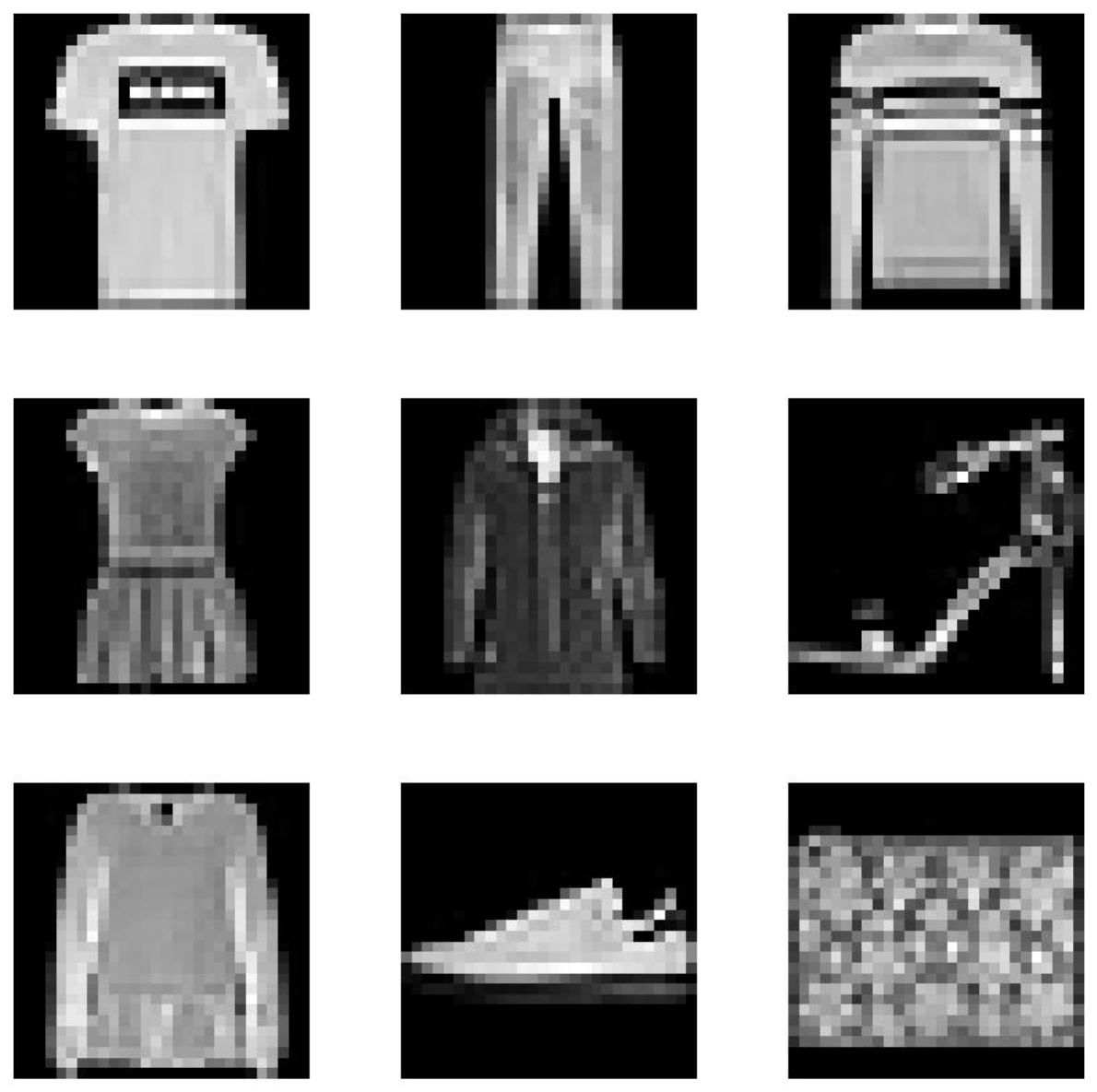}
        \caption{Fashion-MNIST}
        \label{fig:fashionmnist_sample}
    \end{subfigure}
    \caption{Sample images from the MNIST, CIFAR-10, and Fashion-MNIST datasets.}
    \label{fig:datasets_samples}
\end{figure}

\subsection{KANs vs MLPs for Image Classification}

In this experiment, we explore the potential of KANs in addressing various challenges in image classification tasks. We begin by evaluating the performance of standalone KAN architectures and subsequently compare their results with MLPs with identical structures.

The KAN models are trained with varying depths \(d\) in the format [input width, units of hidden layer 1, units of hidden layer 2, ... units of hidden layer \textit{d-2}, output size]. For example, for the MNIST dataset, the input size is \(28 \times 28 = 784\), and with 10 classes, the KAN architecture could be [784, units, 10]. The 'units' for a given layer represent the number of univariate learnable functions and can be considered analogous to the number of neurons in an MLP layer.

To ensure a fair comparison, the MLP model structures are represented in the same format. All models were trained on their native image resolutions (e.g., 28x28 for MNIST). A fixed learning rate of 0.05 was used across all experiments.

\begin{figure}[h]
    \centering

    % MNIST Accuracy
    \begin{subfigure}[b]{0.44\textwidth}
        \centering
        \includegraphics[width=\textwidth]{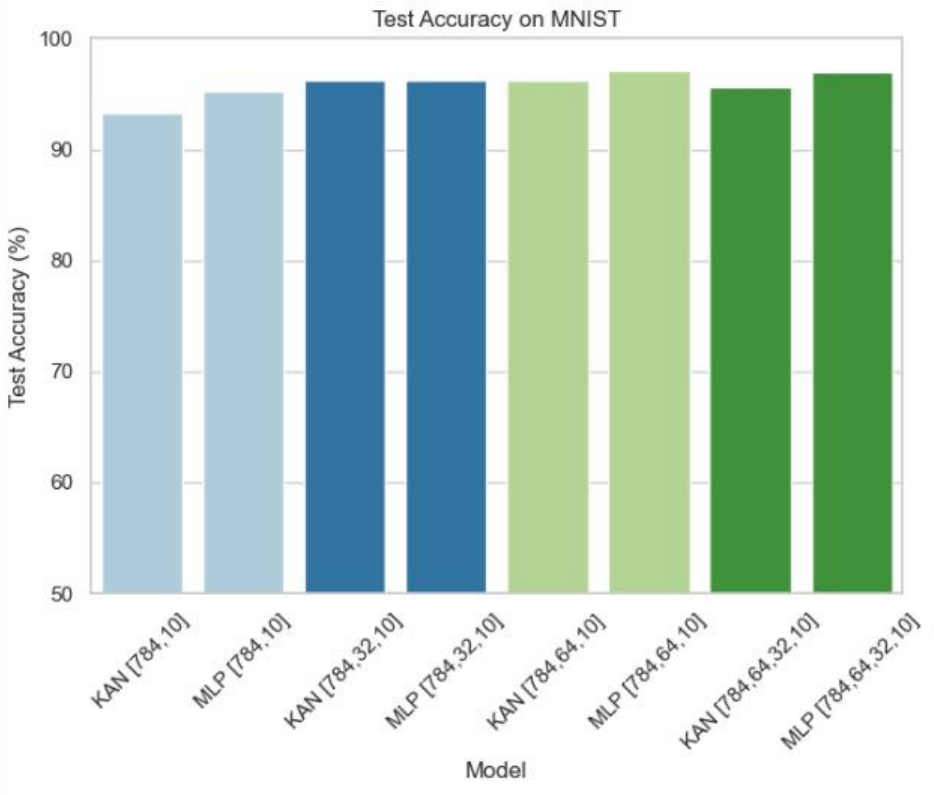}
        \caption{Test Accuracy on MNIST}
        \label{fig:mnist1}
    \end{subfigure}
    \hfill
    % MNIST Parameters
    \begin{subfigure}[b]{0.45\textwidth}
        \centering
        \includegraphics[width=\textwidth]{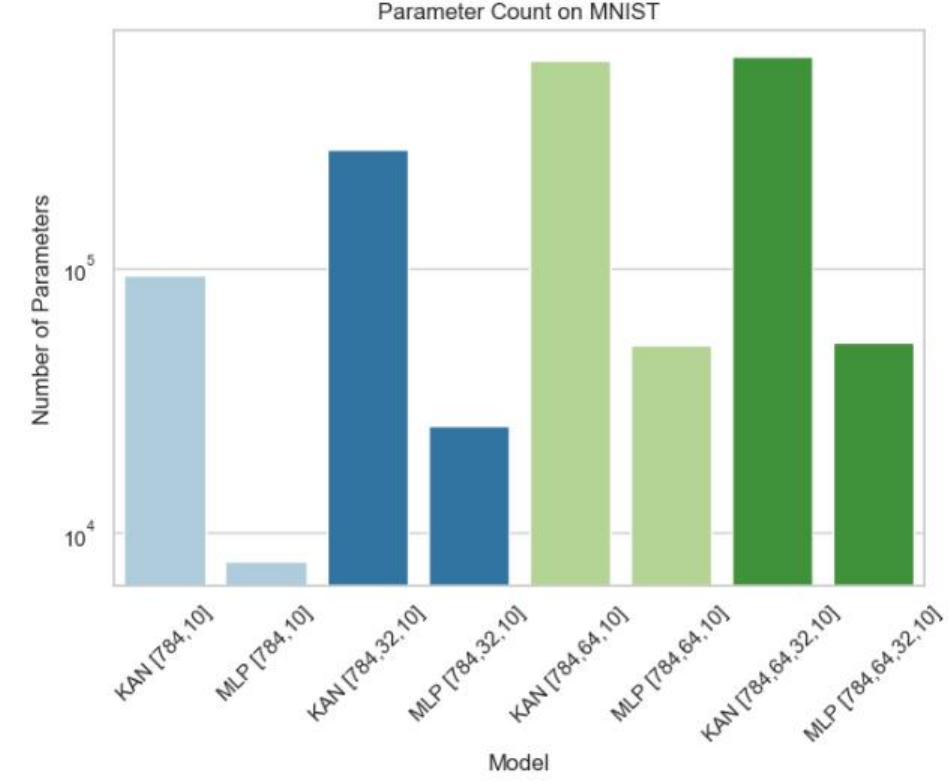}
        \caption{Parameter Count on MNIST}
        \label{fig:mnist2}
    \end{subfigure}

    \vspace{10pt} % Add some vertical space between rows
    
    % CIFAR-10 Accuracy
    \begin{subfigure}[b]{0.45\textwidth}
        \centering
        \includegraphics[width=\textwidth]{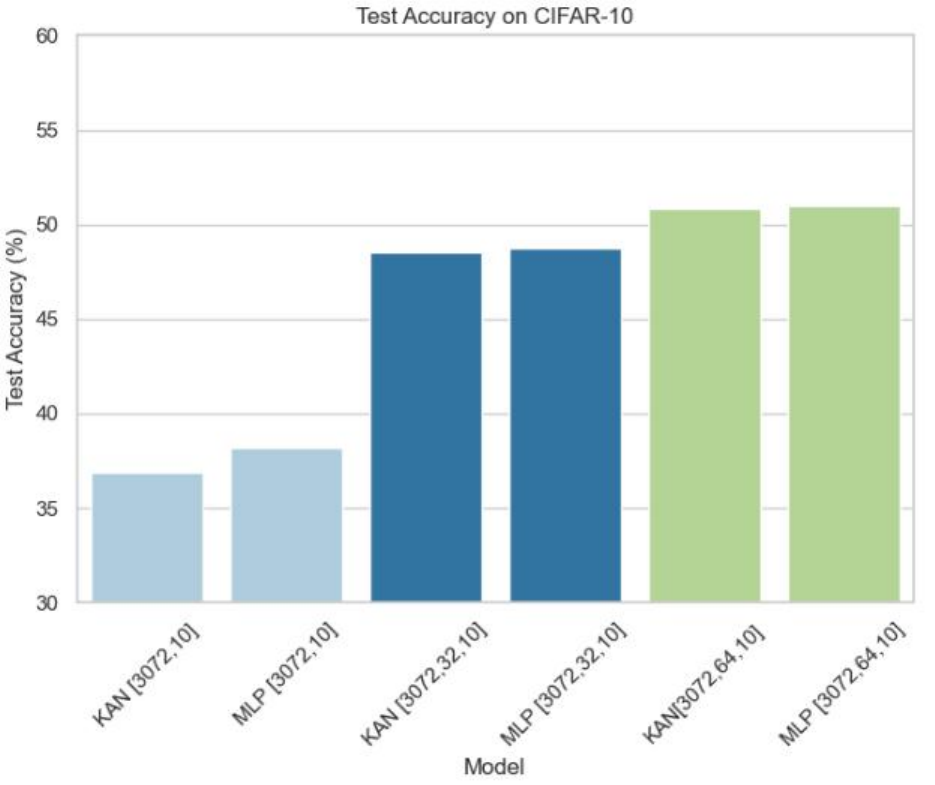}
        \caption{Test Accuracy on CIFAR-10}
        \label{fig:cifar1}
    \end{subfigure}
    \hfill
    % CIFAR-10 Parameters
    \begin{subfigure}[b]{0.45\textwidth}
        \centering
        \includegraphics[width=\textwidth]{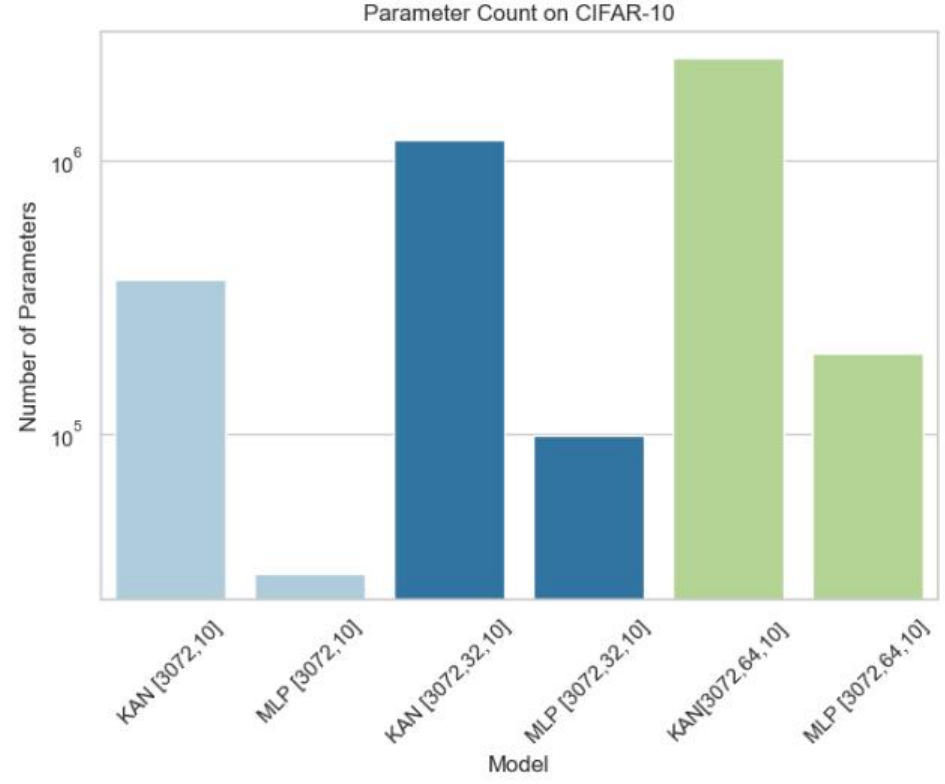}
        \caption{Parameter Count on CIFAR-10}
        \label{fig:cifar2}
    \end{subfigure}
    
    \vspace{10pt} % Add some vertical space between rows
    
    % Fashion-MNIST Accuracy
    \begin{subfigure}[b]{0.455\textwidth}
        \centering
        \includegraphics[width=\textwidth]{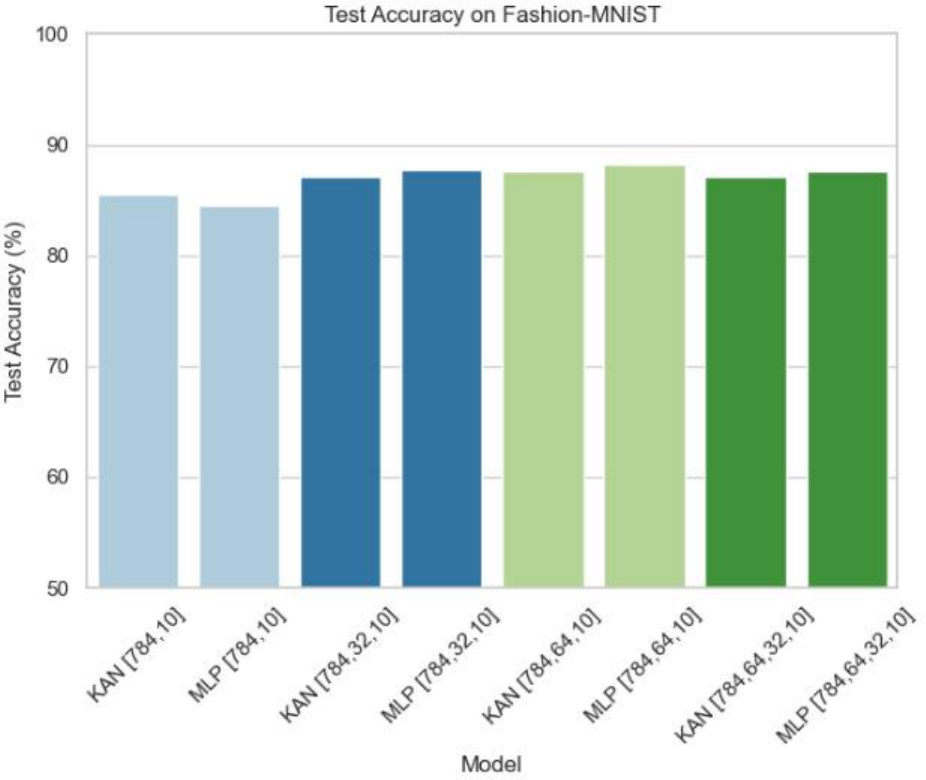}
        \caption{Test Accuracy on Fashion-MNIST}
        \label{fig:fmnist1}
    \end{subfigure}
    \hfill
    % Fashion-MNIST Parameters
    \begin{subfigure}[b]{0.45\textwidth}
        \centering
        \includegraphics[width=\textwidth]{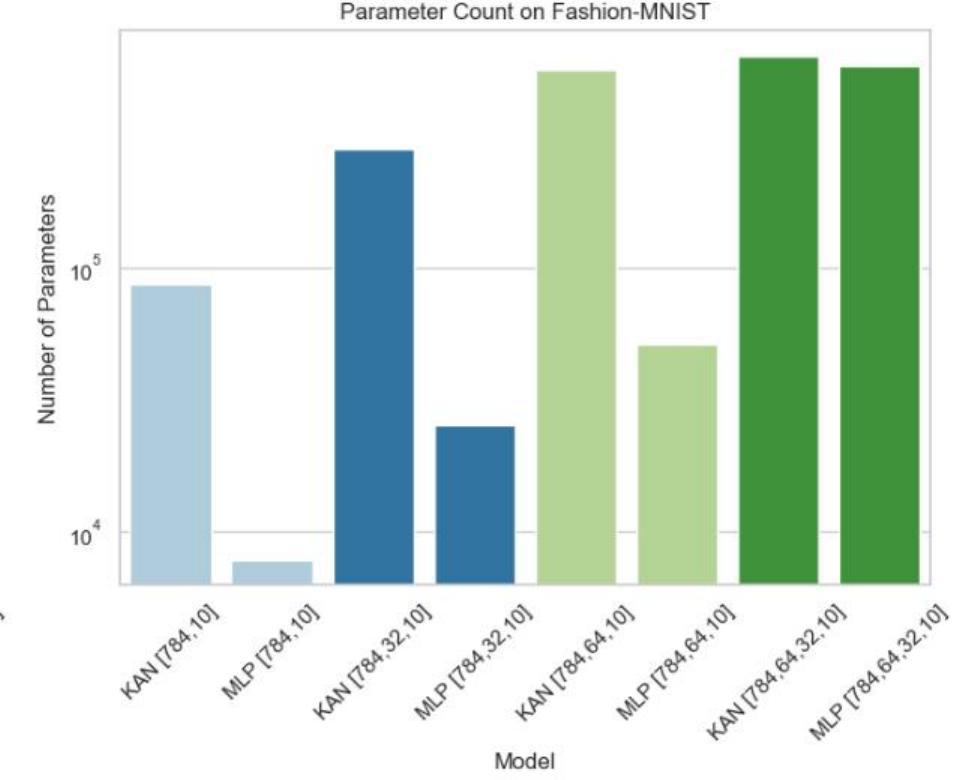}
        \caption{Parameter Count on Fashion-MNIST}
        \label{fig:fmnist2}
    \end{subfigure}
    
    \caption{Comparison of KAN and MLP models across various datasets in terms of test accuracy and parameter count.}
    \label{fig:KANMLP_Comparison}
\end{figure}

Given the complexity of KANs, a comprehensive series of experiments was performed to determine the optimal values for hyperparameters such as order and grid size, which are elaborated in detail in the upcoming sections. The values displayed in Figure \ref{fig:KANMLP_Comparison} represent the best accuracy achieved with these optimizations. The parameter counts shown also correspond to the KAN model with these optimal hyperparameter choices.

From our experiments, it becomes evident that while MLPs and KANs demonstrate nearly equivalent performance when using identical architectures, MLPs generally exhibit superior parameter efficiency, requiring fewer parameters to achieve similar or better accuracy. Moreover, the high parameter count of KAN models imposes substantial computational and memory burdens, making them difficult to train on standard hardware. This, coupled with the sensitivity of KANs to hyperparameter tuning, suggests that while KANs offer a unique approach to neural network architecture, their practical application in image classification tasks is limited by these significant challenges. MLPs emerge as the more optimal choice for image classification tasks, especially when computational efficiency, parameter count, and ease of training are prioritized.

\subsection{Hyperparameters Sensitivity Analysis}
In this section, we analyze the impact of two critical hyperparameters in KANs—grid and order—on model accuracy and parameter count. The experiments were conducted using two KAN architectures, KAN [784,64,10] and KAN [784,32,10], applied to the MNIST dataset.

For each architecture, we systematically varied the grid and order values to assess their influence on accuracy and the number of parameters. The definitions of these hyperparameters are as follows:
\begin{itemize}
    \item \textbf{Grid}: Refers to the discretization level of the univariate functions on the network edges, with finer grids providing more detailed representations.
    \item \textbf{Order}: Denotes the depth of the network, corresponding to the number of layers composing these univariate functions.
\end{itemize}

\begin{figure}[h]
    \centering
    \begin{subfigure}[b]{0.94\textwidth}
        \centering
        \includegraphics[width=\textwidth]{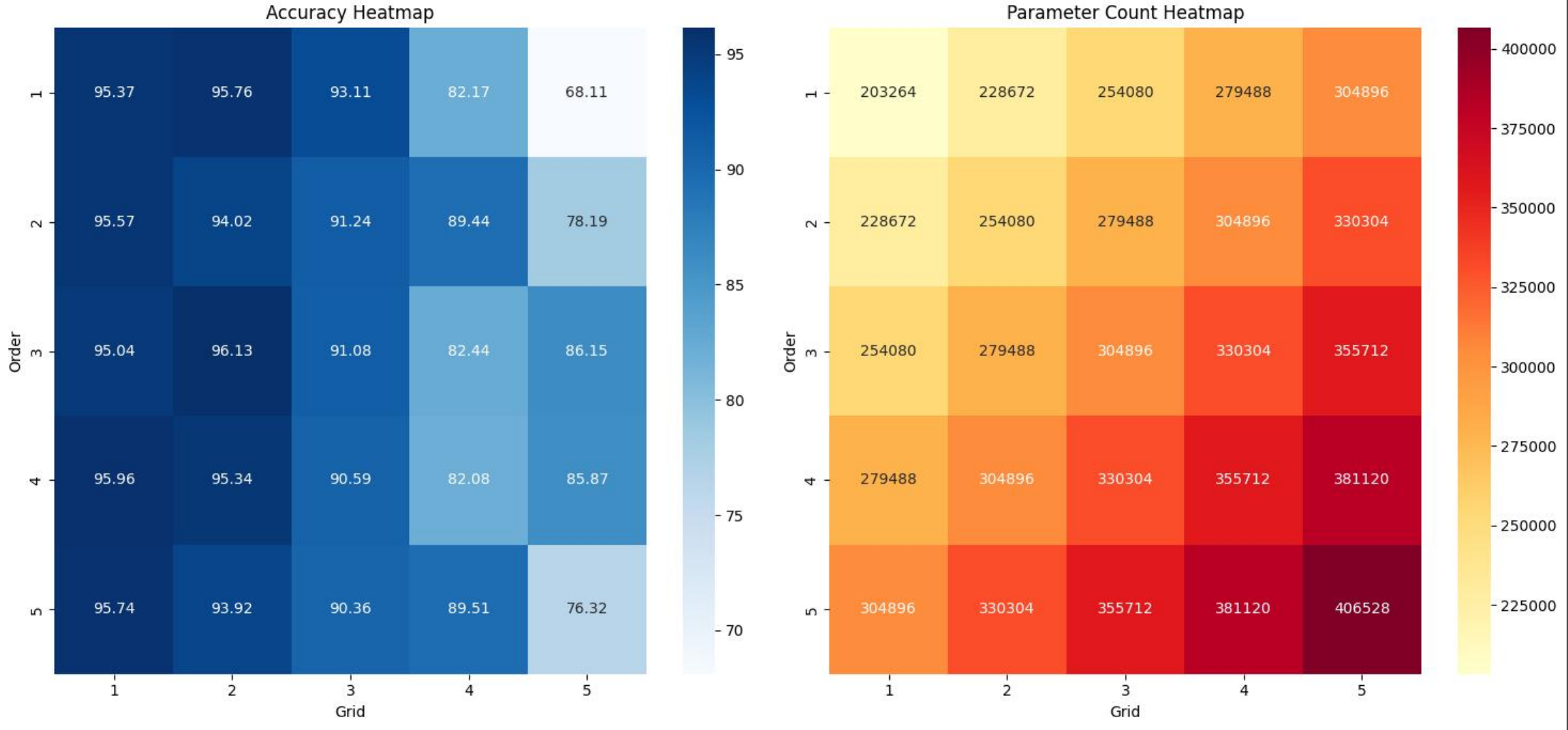}
        \caption{KAN [784,32,10]}
        \label{fig:ablation2}
    \end{subfigure}
    \hfill
    \begin{subfigure}[b]{0.94\textwidth}
        \centering
        \includegraphics[width=\textwidth]{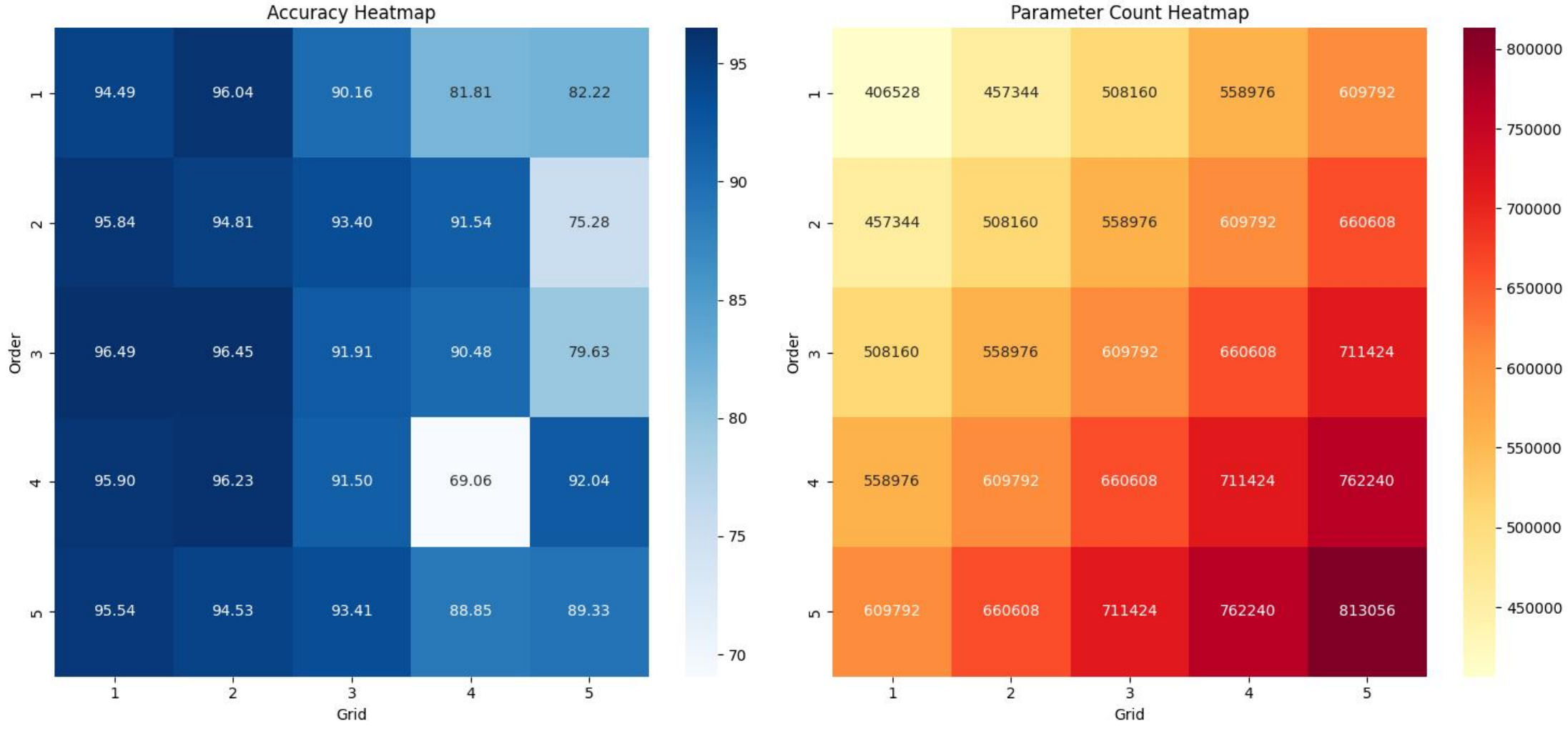}
        \caption{KAN [784,64,10]}
        \label{fig:ablation1}
    \end{subfigure}
    \caption{Heatmaps showing the impact of grid and order on accuracy and parameter count for two KAN architectures on MNIST.}
    \label{fig:heatmap}
\end{figure}

The results, represented as heatmaps in Figure~\ref{fig:heatmap}, provide a clear visualization of the relationship between grid/order values, accuracy, and parameter count.
From the heatmaps, several key insights emerge. Regarding accuracy, moderate grid and order values generally yield the highest accuracy. The best performance was observed around Grid 2 to 3 with lower order values (such as 1 or 2). While increasing the grid value typically enhances accuracy, this improvement only occurs up to a certain point; beyond this, the benefits diminish or even reverse. This trend is particularly noticeable when higher orders are combined with very fine grids, which often leads to overfitting and a consequent drop in accuracy. Additionally, higher order values tend to reduce accuracy when paired with larger grids, indicating that overly complex models struggle with generalization.

As expected, the parameter count increases linearly with both grid and order values. However, it is important to note that extreme values for grid and order (e.g., greater than or equal to 10) were not explored in our experiments. These configurations are computationally expensive, leading to significantly longer training times with no corresponding improvement in performance, and in some cases, even worse outcomes. Therefore, they were deemed not worth investigating.

Our experiments suggest that the optimal configuration for KANs, in terms of grid and order, lies in a balanced region where neither parameter is too low nor too high. This balanced approach maximizes model performance while maintaining computational efficiency.

\subsection{Variation of KANs and Hybrid Models}
Despite standalone KANs not being ideal for image classification tasks, we believe that it could be worth experimenting with variations of KAN independently as well as in conjunction with other networks such as CNNs.

\subsubsection{Efficient KANs - More efficient implementation}
The work on EfficientKAN \cite{efficient} proposes a more optimized implementation of KANs, improving efficiency by leveraging the fact that activation functions are linear combinations of a fixed set of basis functions. Mathematically, if $\phi(x)$ is the activation function, it can be written as:
\begin{equation}
\phi(x) = \sum_{k} c_k B_k(x)
\end{equation}

where $B_k(x)$ are the B-spline basis functions, and $c_k$ are the coefficients. In the original implementation, for a layer with \textit{i} inputs and \textit{o} outputs, the input is expanded to a tensor of shape (\textit{b,o,i}), where \textit{b} is the batch size. This expansion causes significant computational overhead for KANs. In EfficientKAN, instead of expanding the input tensor, the input is directly activated using the basis functions $B_k(x)$. For an input tensor $X$ of shape $(\text{batch\_size}, \text{in\_features})$, we compute:
\begin{equation}
\Phi(X) = [B_1(X), B_2(X), \ldots, B_m(X)]
\end{equation}

where $\Phi(X)$ is now a tensor where each element has been passed through each basis function $B_k$. The outputs from the basis functions are then linearly combined using the coefficients $c_k$. This can be efficiently performed using matrix multiplication. The final output for each layer can be expressed as:
\begin{equation}
Y = \Phi(X) C 
\end{equation}

where $C$ is the matrix of coefficients $c_k$. This reformulation simplifies the computation process to matrix multiplication, thus enhancing both forward and backward passes. This change also allows KANs to handle images with higher resolutions.
% such as \textit{28x28}.

We experimented with EfficientKAN for similar tasks as shown in Table \ref{tab:2}, comparing test accuracies on the MNIST and CIFAR-10 datasets. The experiments were conducted on images of size ${28\times28}$ for 10 epochs with a constant learning rate of 0.01.

\begin{table}[h]
\centering
\begin{tabular}{lccccc}
\toprule
Model & Test Accuracy (MNIST)\\
\midrule
EfficientKAN [784, 16, 10] & 0.945  \\
EfficientKAN [784, 32, 10] & 0.962   \\
EfficientKAN [784, 64, 10] & 0.973   \\
MLP [784, 16, 10] & 0.942 \\
MLP [784, 32, 10] & 0.966   \\
MLP [784, 64, 10] & 0.973   \\
\bottomrule
\end{tabular}
\caption{Comparison of EfficientKAN and MLP architectures for image classification on the MNIST and CIFAR-10 datasets.}
\label{tab:2}
\end{table}

The results in Table \ref{tab:2} show that KANs and MLPs perform very similarly for similar structures. We note that when directly passing flattened images as inputs, KANs have more parameters due to the utilization of splines. This arises because, for a network with depth $L$, layers of equal width $N$, and spline of order $k$ and the number of grids $G$, the parameters of KAN are on the scale of $O(N^2L(G+K))$, whereas for an MLP, it is $O(N^2L)$. However, due to the suggestions of the original KAN paper that KANs require smaller values of \textit{N} and our previous observations of KANs' unsuitability for directly working with spatial data, we believe KANs could be better utilized in the latent space. We experimented with the latent space generated after flattening the features extracted by a convolutional layer, replacing the Fully Connected (FC) Layers with EfficientKAN implementations.

\subsubsection{Convolutional KANs}
The Convolutional KAN layers \cite{convkan} build upon the EfficientKAN implementation by incorporating non-linear activation functions of KANs into traditional convolutional layers. KAN Convolutions are similar to standard convolutions but utilize a learnable non-linear activation function at each element, which is then summed. The kernel of the KAN Convolution is equivalent to a KAN Linear Layer with four inputs and one output neuron. For each input \(i\), a learnable function \(\phi_i\) is applied, and the resulting pixel of that convolution step is the sum of \(\phi_i(x_i)\). Here, \textit{K} refers to the kernel used in the standard convolution operation.
A simple example for computation with convolution and KAN convolution is provided below:
\[
K = \begin{bmatrix}
k_{11} & k_{12} \\
k_{21} & k_{22}
\end{bmatrix}
\]

\[
\phi = \begin{bmatrix}
\phi_{11} & \phi_{12} \\
\phi_{21} & \phi_{22}
\end{bmatrix}
\]

\[
\text{Input Image} = \begin{bmatrix}
a_{11} & a_{12} & \cdots & a_{1m} \\
a_{21} & a_{22} & \cdots & a_{2m} \\
\vdots & \vdots & \ddots & \vdots \\
a_{n1} & a_{n2} & \cdots & a_{nm}
\end{bmatrix}
\]

\[
\text{Output of Standard Convolution} = \begin{bmatrix}
k_{11}a_{11} & k_{12}a_{12} & k_{21}a_{13} & k_{22}a_{14} & \cdots \\
k_{11}a_{21} & k_{12}a_{22} & k_{21}a_{23} & k_{22}a_{24} & \cdots \\
\vdots & \vdots & \ddots & \vdots \\
\end{bmatrix}
\]

\[
\text{Output of KAN Convolution} = \begin{bmatrix}
\phi_{11}a_{11} & \phi_{12}a_{12} & \phi_{21}a_{13} & \phi_{22}a_{14} & \cdots \\
\phi_{11}a_{21} & \phi_{12}a_{22} & \phi_{21}a_{23} & \phi_{22}a_{24} & \cdots \\
\vdots & \vdots & \ddots & \vdots \\
\end{bmatrix}
\]

\begin{table}[h]
  \centering
  \caption{Comparison of Convolutional KAN with other models}
  \begin{tabular}{ccc}
    \hline
    Model & Parameter Count & Test Accuracy \\
    \hline
    Standard CNN (Small) & 34k & 97.9\\
 Standard CNN (Medium)& 157k&99.10\\
 Convolutional KAN with 1 layer MLP & 7.4k&98.53\\
    Convolutional KAN with 2 layer MLP& 163.7k & 98.58\\
    Convolutional KAN with KAN & 94.2k & 98.90\\
    Regular CNN + KAN & 95k& 98.75\\
    \hline
  \end{tabular}
  \label{tab:convkan}
\end{table}
We conducted experiments using KAN Convolution layers and standard Convolutional Neural Networks, as well as exploring various hybrid models involving combinations of MLPs and KANs. We evaluated their performance on the MNIST dataset. The results, presented in Table \ref{tab:convkan}, indicate that Convolutional KANs can achieve higher accuracy compared to standard CNNs and hybrid models, demonstrating the potential of integrating KANs into convolutional architectures.

\section{Conclusion}
In our work, we evaluated the suitability of Kolmogorov Arnold Networks (KANs) for computer vision tasks especially the classical image classification problem, focusing on their performance, parameter efficiency, computational challenges, and potential integrations with other models. We compared KANs with Multi-layer Perceptrons (MLPS) for these tasks, which showed that KANs in their standalone form, offer comparable accuracy to MLPs, but come at a significantly higher parameter count and computational cost. Additionally, factors like high sensitivity to hyperparameters like grid and order are also downsides to KAN's applicability. We also explored implementation-wise modifications to KAN like EfficientKAN which addresses some of these aforementioned concerns by reformulating the computational steps and Convolutional KAN, where learnable activations are embedded into the convolutional layers. Overall, while KANs are not yet a strong alternative for image classification due to their excessive computational overhead, their use in conjunction with other architectures is a promising avenue for future research. Future works can focus on refining these architectures, exploring their scalability and applications on more complex problems.


\begin{thebibliography}{9}

\bibitem{b1} 
McCulloch, W. S., \& Pitts, W. (1943). A logical calculus of the ideas immanent in nervous activity. The bulletin of mathematical biophysics, 5, 115-133. 

\bibitem{b2} 
Rumelhart, D. E., Hinton, G. E., \& Williams, R. J. (1986). Learning representations by back-propagating errors. nature, 323(6088), 533-536.

\bibitem{b3}
LeCun, Y., \& Bengio, Y. (1995). Convolutional networks for images, speech, and time series. The handbook of brain theory and neural networks, 3361(10), 1995. 
\bibitem{b4}
Dosovitskiy, A. (2020). An image is worth 16x16 words: Transformers for image recognition at scale. arXiv preprint arXiv:2010.11929. 
\bibitem{b5}
Krizhevsky, A., Sutskever, I., \& Hinton, G. E. (2012). Imagenet classification with deep convolutional neural networks. Advances in neural information processing systems, 25.
% \bibitem{b6}
% Simonyan, Karen, and Andrew Zisserman. "Very deep convolutional networks for large-scale image recognition." \textit{arXiv preprint arXiv:1409.1556} (2014). 
% \bibitem{b7}
% He, Kaiming, et al. "Deep residual learning for image recognition." \textit{Proceedings of the IEEE conference on computer vision and pattern recognition}. 2016.  
\bibitem{image}LeCun, Y., Boser, B., Denker, J. S., Henderson, D., Howard, R. E., Hubbard, W., \& Jackel, L. D. (1989). Backpropagation applied to handwritten zip code recognition. Neural computation, 1(4), 541-551.
\bibitem{object} Girshick, R., Donahue, J., Darrell, T., \& Malik, J. (2014). Rich feature hierarchies for accurate object detection and semantic segmentation. In \textit{Proceedings of the IEEE conference on computer vision and pattern recognition} (pp. 580-587). 
\bibitem{segment}Long, J., Shelhamer, E., \& Darrell, T. (2015). Fully convolutional networks for semantic segmentation. In Proceedings of the IEEE conference on computer vision and pattern recognition (pp. 3431-3440).
\bibitem{industry}Abiodun, O. I., Jantan, A., Omolara, A. E., Dada, K. V., Mohamed, N. A., \& Arshad, H. (2018). State-of-the-art in artificial neural network applications: A survey. \textit{Heliyon}, \textit{4}(11). 
\bibitem{healthcare} Shahid, N., Rappon, T., \& Berta, W. (2019). Applications of artificial neural networks in health care organizational decision-making: A scoping review. PloS one, 14(2), e0212356.
\bibitem{hyp}Pourkamali-Anaraki, F. (2024). Kolmogorov-arnold networks in low-data regimes: A comparative study with multilayer perceptrons. \textit{arXiv preprint arXiv:2409.10463}. 
\bibitem{security}Dixit, P., \& Silakari, S. (2021). Deep learning algorithms for cybersecurity applications: A technological and status review. Computer Science Review, 39, 100317..
\bibitem{auto}De Brabandere, B., Neven, D., \& Van Gool, L. (2017). Semantic instance segmentation for autonomous driving. In Proceedings of the IEEE Conference on Computer Vision and Pattern Recognition Workshops (pp. 7-9).
\bibitem{relu} Nair, V., \& Hinton, G. E. (2010). Rectified linear units improve restricted boltzmann machines. In Proceedings of the 27th international conference on machine learning (ICML-10) (pp. 807-814).
\bibitem{swish}Ramachandran, P., Zoph, B., \& Le, Q. V. (2017). Searching for activation functions. arXiv preprint arXiv:1710.05941.

\bibitem{b8}Liu, Z., Wang, Y., Vaidya, S., Ruehle, F., Halverson, J., Soljačić, M., ... \& Tegmark, M. (2024). Kan: Kolmogorov-arnold networks. \textit{arXiv preprint arXiv:2404.19756}. 
\bibitem{mnist}LeCun, Y., Bottou, L., Bengio, Y., \& Haffner, P. (1998). Gradient-based learning applied to document recognition. Proceedings of the IEEE, 86(11), 2278-2324. 
\bibitem{cifar}
Krizhevsky, A., \& Hinton, G. (2009). Learning multiple layers of features from tiny images. 
\bibitem{fmnist}Xiao, H., Rasul, K., \& Vollgraf, R. (2017). Fashion-mnist: a novel image dataset for benchmarking machine learning algorithms. \textit{arXiv preprint arXiv:1708.07747}. 
\bibitem{adv}Hou, Y., \& Zhang, D. (2024). A comprehensive survey on kolmogorov arnold networks (kan). \textit{arXiv preprint arXiv:2407.11075}. 
Xiao, H., Rasul, K., \& Vollgraf, R. (2017). Fashion-mnist: a novel image dataset for benchmarking machine learning algorithms. arXiv preprint arXiv:1708.07747. 
\bibitem{efficient}
Efficient-KAN. https://github.com/Blealtan/efficient-kan.git, 2024 
\bibitem{neural}Rumelhart, D. E., Hinton, G. E., \& Williams, R. J. (1986). Learning representations by back-propagating errors. nature, 323(6088), 533-536.
\bibitem{pattern} Bishop, C. M. (1995). Neural networks for pattern recognition. Clarendon Press google schola, 2, 223-228.
\bibitem{convkan} Bodner, A. D., Tepsich, A. S., Spolski, J. N., \& Pourteau, S. (2024). Convolutional Kolmogorov-Arnold Networks. \textit{arXiv preprint arXiv:2406.13155}. 
\bibitem{superior}Yu, R., Yu, W., \& Wang, X. (2024). Kan or mlp: A fairer comparison. arXiv preprint arXiv:2407.16674.
\bibitem{kar-theorem}Kolmogorov, A. N. (1961). On the representation of continuous functions of several variables by superpositions of continuous functions of a smaller number of variables. American Mathematical Society.
\bibitem{survey}Somvanshi, S., Javed, S. A., Islam, M. M., Pandit, D., \& Das, S. (2024). A Survey on Kolmogorov-Arnold Network. \textit{arXiv preprint arXiv:2411.06078}. 
\bibitem{bsplines}De Boor, C. (1972). On calculating with B-splines. Journal of Approximation theory, 6(1), 50-62.
\bibitem{cnnfeature}Krizhevsky, A., Sutskever, I., \& Hinton, G. E. (2012). Imagenet classification with deep convolutional neural networks. Advances in neural information processing systems, 25.

\end{thebibliography}
\end{document}